%% file: main.tex
\pdfoutput=1

\documentclass[11pt]{article}
\newcommand{\method}[1]{\textsc{#1}}
\usepackage[final]{acl}
\usepackage{makecell}
\usepackage{adjustbox}
\usepackage{xcolor}
\usepackage{float}
\usepackage{array}
\usepackage{placeins} 
\usepackage{multirow}
\usepackage{graphicx}
\usepackage{booktabs}
\usepackage{geometry}
\usepackage{comment}
\usepackage{adjustbox} 
\usepackage{lscape}    

\usepackage{times}
\usepackage{amsmath}
\usepackage{amssymb}
\usepackage{arydshln}
\usepackage{hyperref}
\usepackage{enumitem}
\usepackage{wrapfig}
\usepackage{microtype}
\usepackage{graphicx}
\usepackage{subfigure}
\usepackage{booktabs} 
\usepackage{arydshln}
\usepackage{graphicx}
\usepackage{array}
\usepackage{booktabs}
\usepackage{makecell}
\usepackage{multirow}
\usepackage{bbding}
\usepackage{pifont}
\usepackage{latexsym}
\newcommand{\github}{\raisebox{-1.5pt}{\includegraphics[height=1em]{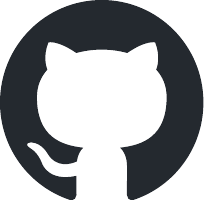}}}
\usepackage[T1]{fontenc}

\usepackage[utf8]{inputenc}

\usepackage{microtype}

\usepackage{inconsolata}

\usepackage{graphicx}
\definecolor{case_1}{rgb}{0.659, 0.855, 0.859}
\definecolor{case_2}{rgb}{0.991, 0.906, 0.835}
\definecolor{case_3}{rgb}{0.5647, 0.7451, 0.8784}
\definecolor{case_4}{rgb}{0.1294, 0.6196, 0.7373}
\definecolor{case_5}{rgb}{0.451, 0.729, 0.447}

\title{DiaLLMs: EHR Enhanced Clinical Conversational System for Clinical Test Recommendation and Diagnosis Prediction}

\author{
  Weijieying Ren$^{\dag}$, 
  Tianxiang Zhao$^{\dag}$, 
  Lei Wang$^{\ddag}$, 
  Tianchun Wang$^{\dag}$, 
  Vasant Honavar$^{\dag}$ \\
  $^{\dag}$ College of Information Sciences and Technology, Pennsylvania State University \\
  $^{\ddag}$ Salesforce AI Research \\
  \texttt{lemontreejying@gmail.com, tkz5084@psu.edu, vuh14@psu.edu}\\
 \hspace*{-0.8cm}\github{} \textbf{\method{DiaLLMs}:} \texttt{ \url{https://github.com/WeijieyingRen/DiaLLMs}} \\
\thanks{Correspondence to Vasant Honavar and Tianxiang Zhao. Feel free to send emails to lemontreejying@gmail.com, tkz5084@psu.edu or vuh14@psu.edu if you have questions.}
}

\begin{document}
\maketitle
\begin{abstract}
Recent advances in Large Language Models (LLMs) have led to remarkable progresses in medical consultation.
However, existing medical LLMs overlook the essential role of Electronic Health Records (EHR) and focus primarily on diagnosis recommendation, limiting their clinical applicability. We propose DiaLLM, the first medical LLM that integrates heterogeneous EHR data into clinically grounded dialogues, enabling clinical test recommendation, result interpretation, and diagnosis prediction to better align with real-world medical practice. To construct clinically grounded dialogues from EHR, we design a Clinical Test Reference (CTR) strategy that maps each clinical code to its corresponding description and classifies test results as "normal" or "abnormal". Additionally, DiaLLM employs a reinforcement learning framework for evidence acquisition and automated diagnosis. To handle the large action space, we introduce a reject sampling strategy to reduce redundancy and improve exploration efficiency. Furthermore, a confirmation reward and a class-sensitive diagnosis reward are designed to guide accurate diagnosis prediction.
Extensive experimental results demonstrate that DiaLLM outperforms baselines in clinical test recommendation and diagnosis prediction.
\end{abstract}

\input{Sections/introduction}

\input{Sections/related_works}
\input{Sections/method}
\input{Sections/experiments}
\input{Sections/conclusion}
\input{Sections/Limitations}

\bibliography{custom}
\newpage
\appendix
\input{Sections/Appendix}

\end{document}

%% file: Sections/introduction.tex
\section{Introduction}
Rapid advancements in LLMs \cite{touvron2023llama,wu2023brief} expanded opportunities to improve diagnostic assistance and patient interactions in the healthcare domain\cite{biswas2023role, li2023chatdoctor,shah2024accuracy, qin2023read},
and clinical conversational systems \cite{wang2023huatuo, yang2024zhongjing} have emerged as a promising approach to enhance clinical reasoning and assist doctors with diagnosis. However, 
as shown in Table \ref{tab:medical_models},
existing studies primarily rely on synthetic dialogues generated from medical knowledge graph \cite{wang2023huatuo} or open medical Question Answer (QA) \cite{labrak2024biomistral}
and are difficult to work on real-world healthcare settings which use Electronic Health Records (EHR). Furthermore, the diagnosis workflow involves multiple critical sub-tasks, including the inquiry of clinical lab tests and the interpretation of their results before giving the diagnosis results, which are often neglected in current approaches \cite{zhou2023survey, liu2024survey}. 
Consequently, existing explorations in the clinical conversational system are still far from practical healthcare scenarios, primarily due to challenges in understanding EHR and adapting to real-world diagnosis workflows \cite{li2024scoping}.

\input{Figures/intro/ehr}
\input{Tables/Comparision_motivation}
\textbf{How to learn from EHR data?} EHR data is a comprehensive digital record encompassing a patient’s medical history, treatments, test results, and clinical decisions. 
As shown in Figure \ref{Fig:ehr}, EHR captures multiple clinical visits and can be structured as a dialogue for clinical conversational systems. However, its heterogeneity and domain-specific nature pose a fundamental challenge, limiting compatibility with existing LLMs \cite{li2024scoping}.
Unlike general NLP tasks, EHR data includes numerical values, categorical attributes from clinical tests, and domain-specific terminologies such as ICD codes and LOINC lab identifiers \cite{sui2024table, liang2024foundation}. 
These complexities require precise numerical reasoning and contextual understanding.
While LLMs benefit from massive text-based data and are trained using next-token prediction \cite{zhao2023survey}, they exhibit significant weaknesses in understanding numeric and specialized clinical knowledge \cite{sui2024table}. 

\textbf{Which Service should Clinical Conversational System Provide?} 
The second challenge lies in the alignment between LLM-supported conversational system and clinical workflows.
As shown in Table \ref{tab:medical_models}, existing works primarily focus on diagnosis assistance using "symptom-diagnosis" conversational data, which oversimplifies the diagnostic process and limits practical applicability.
In a typical diagnostic scenario, a patient presents symptoms, prompting the clinician to iteratively gather information through inquiries and clinical lab tests.
This process follows a cycle of `evidence acquisition, results interpretation, and diagnosis confirmation', with the patient's medical trajectory information promptly recorded in the EHR data.
Simulating clinicians' evidence acquisition and automated diagnosis process within LLMs still remains unexplored \cite{zhou2023survey}.

\textbf{Motivation of DiaLLM.} 
In this work, we propose a novel conversational agent, named DiaLLM, which provides an EHR-grounded transformation pipeline and explicitly models clinicians' reasoning processes for evidence acquisition and automated diagnosis.

Technically, the EHR-grounded transformation aims to convert EHR data into dialogues that are aligned with common-sense knowledge and interpretable to LLMs.
It first converts the EHR data into a single or multi-turn dialogue dataset, structured according to the patient's clinical visit timeline. Then, we translate the heterogeneous dialogue into clinically-grounded text by designing a Clinical Test Reference (CTR) strategy.
The CTR facilitates
(1) the translation of standardized clinical codes (such as ICD-9/10, LOINC) into clinically-grounded text; and (2) the interpretation of clinical test results, conditioned on the patient's age and gender.

On top of transformed EHR data, DiaLLM models evidence acquisition and automated diagnosis with a reinforcement learning framework. 
A policy network selects clinical tests (i.e., take actions) or terminates the process to make a diagnosis. 
Upon termination, a supervised classification model is invoked for disease diagnosis.
To handle the large action space of clinical tests, we introduce a novel rejection sampling strategy \cite{bardenet2014towards,fan2023dpok,mandel2016offline} that pre-filters redundant or unnecessary tests, ensuring only clinically relevant ones are selected. 
Meanwhile, since patients can have multiple diagnoses and disease distributions exhibit a long-tail pattern across populations, we propose a new confirmation reward and a class-sensitive classification reward to enhance diagnosis prediction.

\textbf{Evaluation.}
We propose a comprehensive evaluation framework for medical LLMs, assessing both single-turn and multi-turn consultations. Experimental results show that DiaLLM outperforms both general-purpose and medical-specific LLMs in clinical test selection and diagnosis prediction. Our ablation study reveals that integrating the EHR-grounded transformation pipeline and specialized reward modeling significantly improves clinical test comprehensiveness, result interpretation, and early diagnosis accuracy.

%% file: Figures/intro/ehr.tex
\begin{figure}
\centering
\includegraphics[width=0.45\textwidth]{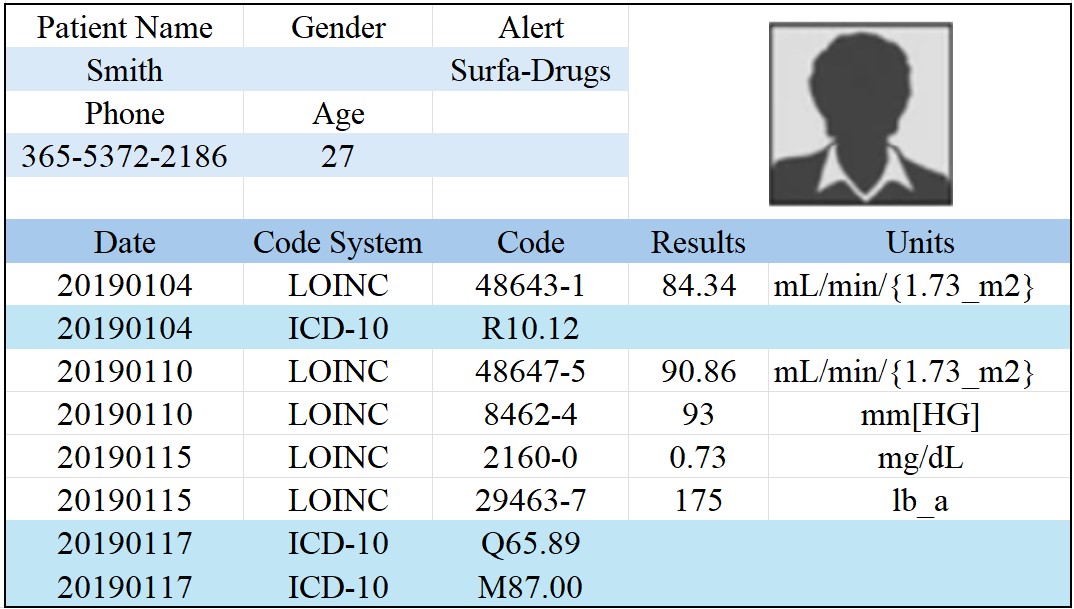}
\caption{ Structured representation of EHR data, illustrating patient demographics, symptoms, clinical tests (e.g., lab test, vital signs), and physician decisions over time.
Symptoms and diagnoses are encoded using ICD codes, while clinical tests follow the LOINC system.}
\label{Fig:ehr}
\vspace{-0.55cm}
\end{figure}

%% file: Tables/Comparision_motivation.tex
\begin{table*}[t]
\centering
\begin{adjustbox}{max width=\linewidth}
\begin{tabular}{ccccc}
\toprule
\textbf{Model} & \textbf{Main Data Source} & \textbf{Function} & \textbf{Training Method} & \textbf{Language} \\
\midrule
\texttt{ChatDoctor \cite{li2023chatdoctor}} & Medical Consultation Website & Diagnosis Assistance & SFT & English \\
\texttt{DoctorGLM \cite{xiong2023doctorglm}}  & Physician-Patient Dialogues & Diagnosis Assistance & SFT & Chinese \\
\texttt{BenTsao \cite{wang2023huatuo}}  & \makecell{Medical knowledge Graphs} & Diagnosis Assistance & SFT & Chinese \\
\texttt{Zhongjing \cite{yang2024zhongjing}}  & \makecell{Proprietary data, Crawled data} & Diagnosis Assistance & SFT + RLHF & Chinese \\
\texttt{ Biomistral\cite{labrak2024biomistral}} & Medical QA & Diagnosis Assistance &SFT & English\\
\texttt{Meditron \cite{chen2024meditron}} & PubMed articles & Diagnosis Assistance &SFT & English\\
\midrule
DiaLLM & EHR data & \makecell{Diagnosis Assistance, \\ Test Results Analysis, \\ Clinical Test Ordering} & PPO & English \\
\bottomrule
\end{tabular}
\end{adjustbox}
\caption{Comparison of Data Sources, Function, Training Method, and Language across Popular Medical Conversational Models.}
\label{tab:medical_models}
\vspace{-0.5cm}
\end{table*}

%% file: Sections/related_works.tex
\section{Related Works}
\textbf{Representation Learning for EHR data.}
The majority of existing studies enhance medical code representations by incorporating external relational information through medical ontologies \cite{choi2017gram,panigutti2020doctor} and knowledge graphs \cite{jiang2023graphcare,wang2023biobridge}. 
However, these established medical knowledge sources are restricted to specific diseases \cite{si2021deep}.
Recent works have attempted to derive clinical concept embeddings from large-scale medical text corpora \cite{ye2021medretriever}, learning relational graphs through self-supervised learning \cite{yao2024self}, contrastive learning \cite{cai2022hypergraph}, or generating medical concepts directly using LLMs \cite{ma2024clibench}.
However, these approaches often lack clinically grounded annotations, making the learned embeddings incompatible with LLM inputs. 

Another line of research improves clinical prediction performance by learning from clinical test results. 
Existing works derive partition functions to learn from numerical and categorical clinical test results \cite{la2019interpretation}, employing methods such as XGBoost \cite{chen2016xgboost}, additive models \cite{hastie2017generalized}, piecewise linear functions \cite{montomoli2021machine}, and logic-based rule learning \cite{rentablog}. However, these approaches can be susceptible to biases introduced by patient population variations and may not align with established guidelines for interpreting test results. 

\noindent
\textbf{Diagnosis-oriented Conversational System.}
Early research primarily focused on ICU-based temporal EHR data \cite{yoon2019asac, fansi2022towards, qin2024risk, he2022scalable} or specific diagnosis categories \cite{he2022scalable, fansi2022towards}, and proposed to model evidence inquiry and diagnosis as a Markov decision process (MDP) \cite{tang2016inquire}. With the rise of foundation models, clinical conversational systems have been explored by tuning on various medical corpora, including clinical conversations collected from online medical consultation website \cite{li2023chatdoctor}, symptom-diagnosis dialogues \cite{toma2023clinical}, medical question-answering pairs \cite{han2023medalpaca}, and knowledge graph-generated dialogues \cite{yang2024zhongjing, wang2023huatuo}.
However, these curated datasets deviate from real-world data distribution, lack essential clinical test support and do not interpret lab test results. These shortcomings limit their real-world applicability. For a comprehensive review, see \cite{zhou2023survey}.
In contrast, our approach constructs single-turn and multi-turn dialogue data leveraging real-world EHR data, and facilitates lab test requesting and diagnosis prediction.

%% file: Sections/method.tex
\section{Methodology}
\input{Figures/intro/overview}
\subsection{Diagnostic Conversational System Setup}
\textbf{Task Definition and Notations.}  
DiaLLM encapsules the evidence acquisition and diagnosis automation with the following steps: 
\begin{enumerate}
\item \textbf{Initial Query:} The patient provides demographic information $\mathbf{d}$ and symptoms $\mathbf{s}$.  
\item \textbf{Clinical Test Recommendation:} The LLM agent suggests initial clinical tests $\mathbf{c_0}$.
\item \textbf{Test Result Analysis and Follow-ups:} 
Upon receiving the patient's initial clinical test results $\mathbf{v_0}$,  the agent conducts a clinically grounded analysis and iteratively recommends additional follow-up tests $\mathbf{c_{t>0}}$ and interprets new results $\mathbf{v_{t>0}}$ to gather further evidence. This process continues until a conclusive diagnosis is achieved.
\item  \textbf{Diagnosis Prediction:} The system ultimately predicts the diagnosis $\textbf{y}$.
\end{enumerate}

\noindent
\textbf{Problem Formulation.}
This process can be modeled as a Markov Decision Process (MDP) \( \mathcal{M}(\mathcal{S}, \mathcal{A}, \mathcal{R}, \gamma) \), where:  
 $
  \mathcal{S} = \mathcal{S'} \cup \{\mathbf{s}_\bot\}
$
is the state space, with
$ \mathbf{s}_\bot  $
as the terminal state.  
$ \mathcal{A} = \mathcal{A'} \cup \{\mathbf{a}_\bot\}$
is the action space, with $\mathbf{a}_\bot$ as the stop action.
$\gamma$ is the discount factor.   
Each dialogue cEach dialogue consists of at most \( T \) turns, where $T$ represents the maximum number of visits recorded in the EHR data.onsists of at most \( T \) turns, where $T$ represents the maximum number of visits recorded in the EHR data.
At turn \( t \), the state \( \mathbf{s}_t \in \mathcal{S'} \) encodes socio-demographics, acquired evidence and dialogue history \( \mathbf{h}_t \):  
\begin{equation}
    \mathbf{s}_t = \{\mathbf{d}, \mathbf{s}, \mathbf{c_t}, \mathbf{v_t}, \mathbf{h}_t\}.
\end{equation}  
At turn $t$, the agent selects an action $\mathbf{a_t}$, determining whether to request further tests or stop. The clinical test recommendation model $\mathbf{\pi_{\theta}}(\mathbf{a_t} | \mathbf{s_t})$ governs test selection, while the diagnosis prediction model $\mathbf{\pi_{\phi}}(\mathbf{y} | \mathbf{s}_\bot)$ makes the final diagnosis.  

\noindent
\textbf{Overview of DiaLLM.} 
In this work, we present DiaLLM designed to enhance evidence acquisition, result interpretation, and the diagnostic workflow. It consists of two key components:
(1) a transformation strategy that constructs dialogues from heterogeneous EHR data, aligning them to common-sense texts for the ease of understanding by LLMs; and (2) a reinforcement learning framework to empower the base LLM with evidence acquisition and automated diagnosis capabilities, for which we design a reject sampling strategy and several reward signals to facilitate learning. Both components will be detailed later in this section.

\subsection{Dialogue Data Construction from EHR.}
We introduce an EHR transformation strategy to convert structured EHR into clinically grounded dialogues. This process begins by segmenting patient visit records into conversational episodes when clinical visit intervals are within one week. A major challenge lies in representing medical terminologies (e.g., codes and lab test identifiers) and interpreting lab results. While text-driven LLMs excel in natural language understanding and reasoning, they lack direct exposure to structured EHR data. To address this, we manually curate a \textit{Clinical Test Reference} data to translate heterogeneous EHR data in a clinically meaningful manner.

The transformation comprises two components:
(1) Clinical Code Translation that converts specialized medical terms, including symptom codes $\mathbf{s}$, diagnosis codes $\mathbf{y}$, and clinical test codes 
$\mathbf{c_t}$ into clinically grounded and common-sense language that is easily understandable by LLMs.
An example of the clinical code translation data is provided in Table \ref{tab:clinical-description} in the Appendix.
(2) Clinical Test Interpretation that transforms heterogeneous lab test results using domain knowledge, including grounded reference ranges conditioned on gender and age, and classifies test results as 'normal' or 'abnormal'.
An example of the clinical test interpretation data is shown in Table \ref{tab:clinical-reference} in the Appendix.

\noindent\textbf{Details of Clinical Test Reference database.}
For clinical code translation, symptom and diagnosis codes are commonly recorded using the ICD system, such as ICD-9 and ICD-10. To standardize code descriptions, we build a comprehensive codebase that includes ICD-9, ICD-10, and a mapping from ICD-9 to ICD-10 for standardization. 
Clinical tests are generally coded using the \textbf{LOINC system}. 
We annotate 735 clinical test code descriptions (including both lab test and vital sign codes) 
\footnote{LOINC Organization: \url{https://loinc.org/}} which cover most of diseases
to ensure comprehensive coverage and standardization.

For clinical test interpretation, we annotate 262 commonly used clinical tests based on established medical guidelines\footnote{LOINC Organization: \url{https://loinc.org/}, Mayo Clinic Laboratory: \url{https://www.testcatalog.org/show/NAS}, Corewell Health: \url{https://corewellhealth.testcatalog.org/show/LAB299-1}}. This resulted in 1,163 annotations \footnote{Each clinical test may correspond to multiple reference ranges conditioned on patient age and gender. }  categorized as follows:
(1) Normal Ranges that define reference ranges considering age and gender influences.
(2) Critical Values that identifies thresholds indicating life-threatening conditions requiring immediate intervention.
(3) Demographic Variability that includes age- and gender-specific ranges to account for physiological differences.
(4) Units of Measurement that ensures consistency across international and clinical standards.

\noindent\textbf{Example.}
Given the EHR example: "Female, Age 27, LOINC 2160 0.48 mg/dL," the CTR translates it to: "Given the patient demographic information (Age: 27, gender: Female). 
These lab tests show normal results: Creatinine in Serum or Plasma".
\subsection{Learning for Lab Test and Diagnosis.}
On top of transformed EHR data, DiaLLM models lab test acquisition and automated diagnosis using a reinforcement learning framework. At each step, the model will select follow-up lab tests (i.e., take actions) or termination of this process to make a diagnosis. To facilitate learning, we adopt rejection sampling to reduce the complexity.
During fine-tuning with PPO, both a confirmation-based reward and a Class-sensitive Diagnosis reward are utilized.

\noindent\textbf{Rejection Sampling for Lab Test Selection.}
In the presence of cost pressure, the goal of the agent is to balance timely and accurate evidence acquisition with cost-effective feature selection. The vast number of potential lab tests leads to a large action space, making direct reinforcement learning (RL) inefficient. To mitigate this, we introduce a strategy based on rejection sampling \cite{bardenet2014towards,mandel2016offline} which pre-filters redundant or unnecessary lab tests before execution, ensuring that only clinically relevant tests are selected.

At each decision step $t$, the actor model $\pi(a_t|s_t)$ samples a candidate lab test 
$c_t$ from the full action space. Instead of executing the sampled action immediately, we introduce a rejection sampling mechanism that determines whether the test should be conducted based on its informativeness, cost, and redundancy. The probability of accepting a test 
$p_{accept}(c_t|s_t)$ is defined as:
\begin{equation}
\begin{aligned}
   p_{\text{accept}}(c_t \mid s_t) = & \frac{H(y \mid s_t) -H(y \mid s_t, v_t)}{\max_{c_t} H(y \mid s_t) -H(y \mid s_t, v_t)} \\
   & \cdot
   1[c_t \notin C_{\text{prev}}] ,
   \end{aligned}
\end{equation}
where \(H(y \mid s_t, v_t) \) denotes the entropy of predicted diagnosis given the current sate $s_t$ and value $v_t$ of candidate test $c_t$, capturing how much uncertainty is reduced by performing \( c_t \). 
The redundancy filter \( 1[c_t \notin C_{\text{prev}}] \) prevents reordering previously conducted tests.


\noindent\textbf{Confirmation Reward.}
As the conversation progresses, the agent should systematically acquire more evidence to refine its belief in the correct diagnosis while minimizing uncertainty.
Inspired by potential-based reward shaping \cite{hu2020learning}, we formalize the confirmation reward $R_{\text{co}}$ as:
\begin{align*} 
R_{\text{Co}}(s_t, a_t, s_{t+1})  &= 1_{s_{t+1} \neq s_\perp} \cdot \\
& 
\left( CE(\hat{y}_{t+1}, y) - CE(\hat{y}_t, y) \right),
\end{align*}
where $1_{s_{t+1} \neq s_\perp}$ is an indicator that the terminal state has not been reached yet.

\noindent\textbf{Class-sensitive Diagnosis Reward.}
This reward is designed to provide feedback to the agent by evaluating the quality of its final predicted diagnosis with respect to the ground truth diagnosis 
$y$ once the interaction process is completed. To directly optimize diagnosis prediction while addressing the issue of class imbalance \cite{puthiya2014optimizing,elkan2001foundations}, we introduce a weighted classification reward:
\begin{equation}
R_{\text{Cl}}(s_T) = \sum_{y_i \in y} w_{\text{Cl}}(y_i) \cdot \text{CE}(\hat{y}_i, y_i),
\end{equation}
where $w_{\text{Cl}}(y_i) = \frac{1}{p(y_i)}$ assigns a higher weight to rare diagnoses based on the inverse of the class frequency.
This reward adjusts the importance of correctly predicting minority classes, ensuring that the agent focuses on both high-risk and low-risk diagnoses effectively.

%% file: Figures/intro/overview.tex
\begin{figure*}
\centering
\includegraphics[width=1\textwidth]{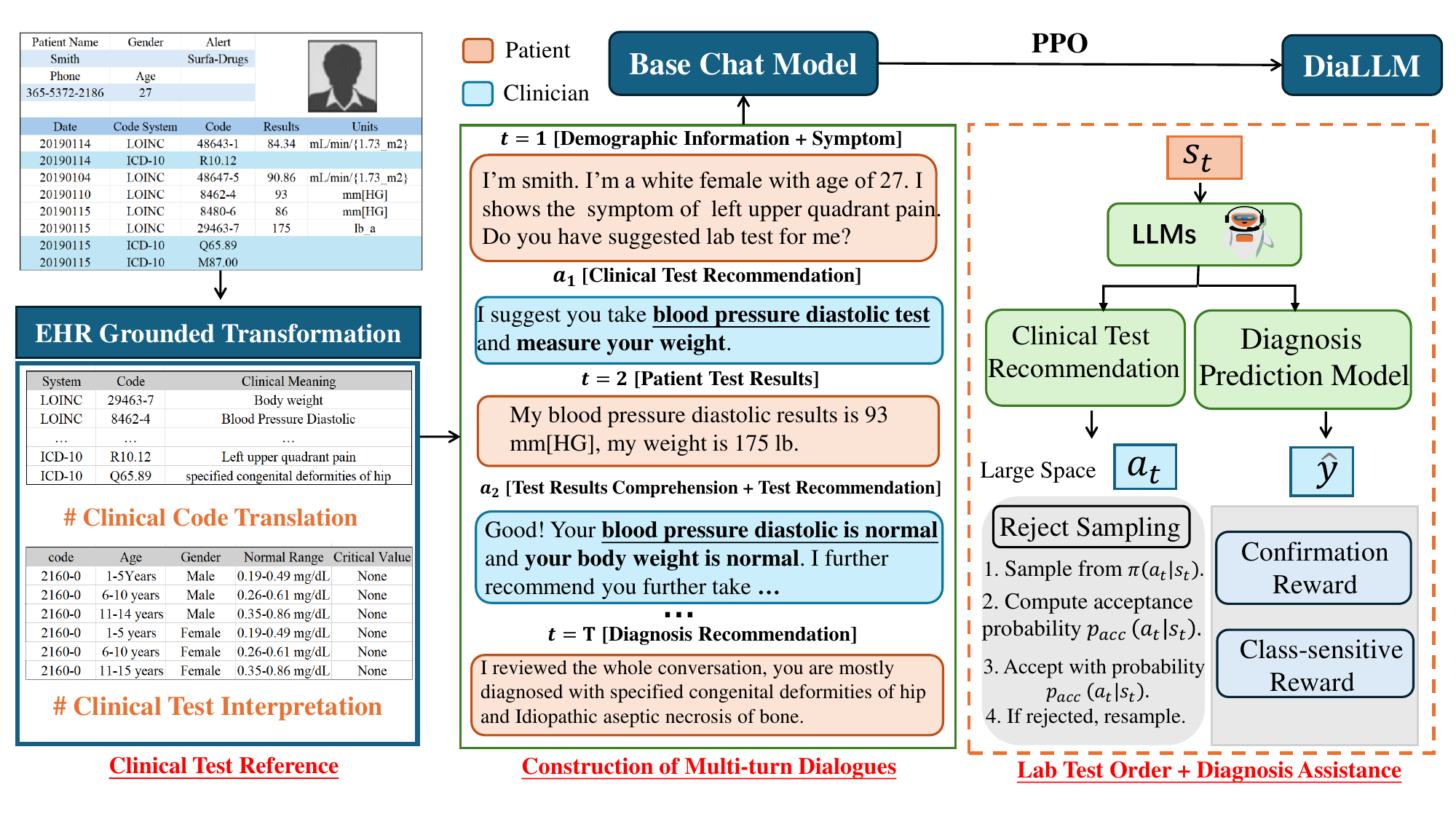}
\caption{DiaLLM operates in two stages: (I) Dialogue Data Construction, where EHR data is transformed into clinically grounded dialogue data, and (II) Reward Modeling, which optimizes clinical test selection and diagnosis prediction.
In Stage I, DiaLLM converts clinical test codes and results into texts using our constructed Clinical Test Reference. In Stage II, we adopt a novel rejection sampling strategy to handle the large action space and incorporate two reward signals to facilitate diagnosis prediction learning.}
\label{Fig:overview}
\vspace{-0.1cm}
\end{figure*}

%% file: Sections/experiments.tex
\section{Experiments}
\subsection{Dataset and Task Description}
To critically evaluate the performance of LLMs on EHR data, we design the dataset with two key considerations: 
(I) \textit{Unexposed}: Ensuring that the dataset has not been previously used by most LLMs. 
(II) \textit{Predictive}: Focusing on diseases that can be predicted solely from EHR data, reflecting their clinical applicability. 
Detailed data statistical analysis is listed in Table \ref{tab:Statistical_Datasets} in Appendix \ref{app::statistic}.

\textbf{NHANES Dataset}. Following \cite{kachuee2019opportunistic}, we construct a diabetes dataset from the public National Health and Nutrition Examination Survey (NHANES) \footnote{https://wwwn.cdc.gov/nchs/nhanes/}. The NHANES dataset does not include clinical visit time information, and we construct a single-turn dialogue based on this data. The dataset includes 8,897 patients and 45 heterogeneous features, such as demographic data, lab results (e.g., total cholesterol, triglycerides), physical measurements (e.g., weight, height), and responses to questionnaires (e.g., smoking, alcohol consumption).

\textbf{TriNetX Dataset}. 
TriNetX\footnote{\href{https://trinetx.com/}{https://trinetx.com/}} is a global health research network providing access to large-scale de-identified patient EHR data. 
TriNetX includes records from 33,105 de-identified patients, spanning data from 1982 to 2023, collected across more than 100 community hospitals and 500 outpatient clinics. 
To satisfy goal (I) and (II), we extract data on metabolic, respiratory, and circulatory diseases, referred to as TriNetX-Metabolic, TriNetX-Respiratory and TriNetX-Circulatory, respectively.


\subsection{Baselines and Evaluation Metrics}
We compare DiaLLM with (1) pretrained foundation models\footnote{For simplicity, we use shortened names later.} including \texttt{Mistralv0.3-7B Instruct} \cite{jiang2023mistral}, \texttt{Llama3.1-8B Instruct} \cite{dubey2024llama}, \texttt{Qwen2.5-7B Instruct} \cite{yang2024qwen2}; and (2) medical-specialized models include  \texttt{BioMistral} \cite{labrak2024biomistral}, 
\texttt{Meditron-7B} \cite{chen2024meditron},
\texttt{Meditron3-8B} \cite{chen2023meditron},
\texttt{Chatdoctor} \cite{li2023chatdoctor}. To evaluate zero-shot performance, we incorporate diagnosis labels in the prompt. For foundation models, we also test an alternative approach where we extract their generated embeddings and train an MLP for prediction.

We evaluate the prediction performance through two dimensions:
(1) Coverage Ability. 
We use the recall@5 and F1 to measure the coverage ability.
Since there are no lab test requests in single-turn dialogues, we evaluate lab test performance on the multi-turn dialogue data using recall@5.
(2) Early-Prediction Ability.
We utilize Mean Reciprocal Rank (MRR) to indicate the effectiveness of early diagnosis prediction.
Details about evaluation metrics are shown in \ref{sec::evaluation_metric} in Appendix. 

\subsection{Implementation Details}
During implementation, we select Llama3.1-8B as the backbone foundation model, and append a mean-pooling layer and a two-layer MLP to it for lab test recommendation and diagnosis prediction. During tuning, $r$ is set to $16$ for the LoRA adapters \cite{hu2021lora}. Batch size is set to $4$ and learning rate is $1e-4$. Training epoch is set to $5$. We randomly run each experiment twice and report the mean. We run experiments on four A100, and train:eval:test is set to 8:1:1.

\subsection{Main Results and Analysis}
\textbf{Results on Single-turn Prediction.}
\input{Tables/single_turn}
For single-turn diagnosis, additional lab test query is not needed. For space limitation, experimental results are presented in Table \ref{tab:single_turn} and Table \ref{tab:NHANES_single_turn} in Appendix \ref{app::NHANES}. 
Several key observations can be made from the following dimensions:
(1) \textbf{Superiority of DiaLLM.}
It is obvious that DiaLLM consistently outperforms all baselines, demonstrating its effectiveness. 
This improvement is mainly due to the EHR transformation pipeline and rewarding models for PPO, which incorporate clinically grounded knowledge, thereby enhancing the model's decision-making abilities.
(2) \textbf{Impact of Tuning.} Comparing base models (e.g., Mistralv0.3-7B, Llama3.1-8B, Qwen2.5-7B) with their supervised fine-tuned counterparts, we observe substantial performance gains across all metrics, hightlighting the importance of task-specific fine-tuning in improving decision-making in clinical dialogue tasks.
(3) \textbf{Performance of clinical-Specific Models.}
The performance of BioMistral and ChatDoctor is significantly lower than that of fine-tuned general-purpose models. This suggests that current biomedical LLMs are primarily trained for general clinical tasks, such as clinical QA, rather than being optimized for real medical data.
(4) \textbf{Comparison with Fine-Tuned Baselines}.
Among fine-tuned models, those on Llama3.1-8B and Qwen2.5-7B perform competitively. 
However, DiaLLM consistently outperforms them, indicating that its enhanced clinical grounding and policy optimization contribute to superior prediction.

\textbf{Results on Multi-turn Prediction.}
\input{Tables/multi_turn}
For multi-turn dialogue data, models can query additional lab tests before giving the final prediction. As shown in Table \ref{tab:multi_turn}, performance of DiaLLM significantly exceeds baseline models, similar to previous single-turn cases. 
The performance of general-used LLMs, e.g., Mistralv0.3-7B, Llama3.1-8B, Qwen2.5-7B still falls short in comparison to fine-tuned models and DiaLLM, indicating that they lack the clinical specificity necessary for multi-turn dialogues.
Medical-specific LLMs like BioMistral, Meditron, and ChatDoctor underperform compared to general-purpose models such as Llama3.1-8B and Qwen2.5-7B. This is likely due to their training on open medical QA datasets or synthetic medical dialogues, which fail to capture the complexity of real-world clinical practice.
\input{Figures/ablation/ablation_single}
\input{Figures/ablation/ablation_multi}

\subsection{Ablation Study}
To evaluate the effectiveness of different components in DiaLLM, we conduct an ablation study to quantify the contributions of the EHR transformation pipeline and the rewards in PPO. For space limitation, experimental results on single-turn and multi-turn dialogue data are shown in Figure \ref{fig:ablation_single}, Figure \ref{fig:ablation_multi} and Figure \ref{Fig:ablation_mhanes} in Appendix \ref{app::aba_nhanes}.

\subsubsection{Ablation Study on EHR Transformation}
We assess the impact of different EHR transformation components by progressively removing them: 
(1) DiaLLM (w/o CT): Removes code translation (CT).
(2) DiaLLM (w/o CTI): Removes clinical test interpretation (CTI).
(3) DiaLLM (w/o Both): Removes all EHR transformation components.

These results highlight the importance of integrating  clinical grounded knowledge into LLMs. 
DiaLLM (w/o Both) results in a noticeable performance drop in both single-turn and multi-turn dialogue data, confirming that the EHR transformation strategy enhances the model's ability to understand EHR data and improves diagnostic predictions. 
Besides, DiaLLM (w/o CTI) exhibits lower performance than DiaLLM (w/o CT) across most datasets, highlighting the necessity of understanding clinical terminology and numerical values.

\subsubsection{Ablation Study on Reward Modeling}
We analyze the influence of specific PPO mechanisms:
(1) DiaLLM (w/o PPO-RJ): Removes reject sampling.
(2) DiaLLM (w/o PPO-Conf): Removes confirmation reward.
(3) DiaLLM (w/o PPO-Imb): Removes imbalanced reward.
Reward modeling is primarily effective for multi-turn dialogue data, and experimental results are shown in Figure \ref{fig:ablation_multi}. 

Each PPO component contributes to model performance, as elaborated in the experimental results. In the TriNetX-Respiratory multi-turn dialogue dataset, removing the imbalanced reward reduces the F1-score to 0.6954, while removing the confirmation reward lowers it to 0.7129. These declines highlight the importance of addressing class imbalance (PPO-Imb) and incorporating confidence-based rewards (PPO-Conf) to improve classification quality, balance predictions, and enhance decision reliability.
Furthermore, the EHR transformation component has a greater impact on model performance than reward modeling in multi-turn dialogue data, showing the importance of aligning medical terms with common-sense knowledge.
\subsection{Case Study}

\noindent
We qualitatively evaluate the diagnostic recommendations of various models using the same example dialogue, as presented in Table \ref{tab::single_turn_prompt} in the Appendix \ref{app:case_study}. 
The selected case presents a complex scenario requiring not only diagnosis prediction but also the integration of additional clinical test evidence.
DiaLLM initially recommended tests for carbon dioxide, chloride, etc. Upon receiving the results, it identified abnormalities in carbon dioxide and chloride levels and suggested further tests for glucose and sodium. Based on these analysis, DiaLLMs can recommend relevant diagnoses.

In contrast, all baseline medical LLMs lacked the ability to inquire about appropriate clinical tests or effectively analyze test results, highlighting a significant gap in diagnostic reasoning. 
BioMistral has a narrow diagnostic scope, potentially missing relevant diagnoses. While Meditron-7B and 8B considered a broader range of conditions, their diagnostic results diverged from the ground truth. ChatDoctor’s diagnoses often lacked relevance to the patient's symptoms and lab results, suggesting limited comprehension of the medical history. Among these, DiaLLM demonstrated the most comprehensive and accurate diagnostic approach, effectively inquiring, analyzing lab results, and adjusting recommendations, showcasing superior practical medical value.

%% file: Tables/single_turn.tex
\begin{table*}[h]
\centering
\resizebox{0.9\textwidth}{!}{
\begin{tabular}{l ccc ccc ccc}
\toprule
\multirow{2}{*}{\textbf{Model}} & \multicolumn{3}{c}{\textbf{TriNetX-Metabolic}} & \multicolumn{3}{c}{\textbf{TriNetX-Respiratory}} & 
\multicolumn{3}{c}{\textbf{TriNetX-Circulatory}} \\  
& Recall@5 & F1 & MRR & Recall@5 & F1 & MRR & Recall@5  & F1 & MRR \\ 
\midrule
Mistralv0.3-7B & $3.64$  & $11.89$  & $10.12$  & $4.54$  & $7.84$  & $12.09$  &  $10.98$ & $8.67$  & $17.57$   \\ 
Llama3.1-8B & $5.34$  & $9.28$  & $10.95$  & $7.49$  & $12.05$  & $15.48$  & $17.02$  &  $10.89$ &  $22.97$   \\ 
Qwen2.5-7B & $8.58$  & $10.51$  & $12.11$  & $10.03$  & $10.64$  & $16.25$ & $5.84$ & $7.82$ & $9.68$   \\ 
\midrule
BioMistral & $39.67$ & $13.82$  & $26.10$ & $25.95$ & $15.50$ & $33.78$ & $20.01$ & $16.62$ & $31.08$   \\ 
ChatDoctor & $19.16$ & $11.57$ & $20.82$  & $35.88$  & $9.71$ & $26.29$ & $10.03$ & $10.03$ & $18.34$   \\ 
Meditron-7B & $6.84$  & $9.31$ & $11.66$  & $11.58$ & $10.34$ & $16.80$ & $15.07$ & $14.71$ & $18.88$  \\ 
Meditron3-8B & $5.99$ & $6.46$ & $10.74$ & $11.98$ & $13.28$ & $15.42$ & $10.13$ & $16.04$ & $19.34$   \\  
\midrule
Mistralv0.3-7B$_{mlp}$ & $66.72$ & $50.92$  & $45.61$  & $67.71$ & $54.95$  & $45.64$  & $67.51$ & $55.95$  & $45.74$  \\ 
Llama3.1-8B$_{mlp}$ & $68.11$ & $55.01$  & $45.76$ & $67.66$ & $56.80$  & $45.77$  & $66.72$  & $56.23$  & $45.76$ \\ 
Qwen2.5-7B$_{mlp}$ & $69.16$ & $52.95$  & $45.72$  & $67.66$  &  $55.21$  & $45.71$  & $68.86$ & $55.18$   & $45.79$  \\ 
\midrule
DiaLLMs & \textbf{79.54}  & \textbf{76.21} & \textbf{47.33} & $\textbf{78.19}$ & $\textbf{74.82}$  & $\textbf{48.18}$  & $\textbf{77.75}$  & $\textbf{74.36}$  & $\textbf{48.14}$ \\ 
\bottomrule
\end{tabular}
}
\caption{Performance Comparison of DiaLLM and Baselines on the Constructed Single-turn Dialogue.}
\label{tab:single_turn}
\end{table*}

%% file: Tables/multi_turn.tex
\begin{table*}[t]
\centering

\setlength\tabcolsep{4pt} 
\resizebox{\textwidth}{!}{
\begin{tabular}{l ccc c| ccc c| ccc c}
\toprule
\multirow{3}{*}{\textbf{Model}} 
& \multicolumn{4}{c|}{\textbf{TriNetX-Metabolic}} 
& \multicolumn{4}{c|}{\textbf{TriNetX-Respiratory}} 
& \multicolumn{4}{c}{\textbf{TriNetX-Circulatory}} \\  
\cmidrule(lr){2-5} \cmidrule(lr){6-9} \cmidrule(lr){10-13}
& \multicolumn{3}{c}{Diagnosis} & Lab Test 
& \multicolumn{3}{c}{Diagnosis} & Lab Test 
& \multicolumn{3}{c}{Diagnosis} & Lab Test \\ 
& Recall@5 & F1 & MRR & Recall@5 
& Recall@5 & F1 & MRR & Recall@5 
& Recall@5 & F1 & MRR & Recall@5 \\ 
\midrule
Mistralv0.3-7B  & $6.89$  & $8.07$  & $13.23$  & $0.58$  
                & $14.44$  & $16.02$  & $11.98$  & $1.96$  
                & $18.00$ & $16.93$  & $25.39$  & $0.21$  \\  
Llama3.1-8B     & $12.44$ &  $11.84$ & $14.79$  & $8.52$  
                & $18.22$  & $17.84$  & $26.08$   & $0.75$
                & $12.72$     & $13.24$      & $21.24$       & $1.87$  \\ 
Qwen2.5-7B      & $7.56$  & $16.68$ & $17.86$  & $0.61$  
                & $5.56$ & $8.90$ & $15.03$  & $1.33$ 
                & $8.22$  & $9.23$  & $15.23$   & $1.75$ \\ 
\midrule
Meditron-7B     & $4.22$  & $8.31$  & $13.28$  & $1.21$  
                & $8.89$  & $13.04$ & $14.67$  & $0.79$  
                & $12.22$  & $12.43$  & $21.90$      & $1.29$\\ 
Meditron3-8B    & $16.22$  & $21.25$  & $21.57$  & $2.90$  
                & $17.33$  & $15.38$  & $24.32$  & $1.37$  
                & $8.22$  & $15.25$   & $17.26$    & $1.91$ \\  
BioMistral      & $9.37$  & $13.54$  & $16.62$  & $2.33$  
                & $12.67$     & $19.51$    & $22.98$    & $1.70$
                & $4.44$     & $13.37$      & $10.56$       & $0.79$  \\ 
Chatdoctor      &  $6.87$ & $9.71$  &  $18.92$ & $1.77$  
                &   $2.89$   & $9.66$    & $9.44$   & $0.95$
                & $16.00$      & $19.22$      & $22.08$       & $1.18$  \\
\midrule
Mistralv0.3-7B$_{mlp}$ & $57.90$ & $56.03$  & $37.42$  & $10.45$  
                & $59.76$      & $59.14$      & $38.90$       & $10.51$  
                &  $59.56$    & $58.90$    & $39.93$   & $10.39$  \\ 
Llama3.1-8B$_{mlp}$ & $57.31$ & $55.92$  & $37.43$  & $10.64$ 
                &   $61.22$   &  $59.58$   &  $39.42$  & $10.97$
               &  $61.21$    & $59.47$    & $40.13$   & $10.16$ \\ 
Qwen2.5-7B$_{mlp}$ & $58.01$ & $56.72$  & $38.37$  & $11.04$ 
                &  $61.91$    & $60.21$    & $39.72$   & $11.13$
                &  $60.89$    & $58.54$    & $38.46$   & $10.30$  \\ 
\midrule
DiaLLMs         
& $\textbf{70.22}$ & $\textbf{73.59}$  & $\textbf{41.88}$  & $\textbf{11.43}$  
& $\textbf{70.89}$  & $\textbf{75.12}$  & $\textbf{40.52}$  & $\textbf{12.31}$  
& $\textbf{71.11}$  & $\textbf{75.39}$  & $\textbf{43.34}$  & $\textbf{11.55}$   \\ 
\bottomrule
\end{tabular}
}
\caption{Performance Comparison of DiaLLM with Baselines on the Constructed Multi-turn Dialogue.}
\label{tab:multi_turn}
\end{table*}

%% file: Figures/ablation/ablation_single.tex
\begin{figure*}[t!]
    \centering
    \subfigure[TriNetX-Metabolic Dataset.]{
        \includegraphics[width=0.3\textwidth]{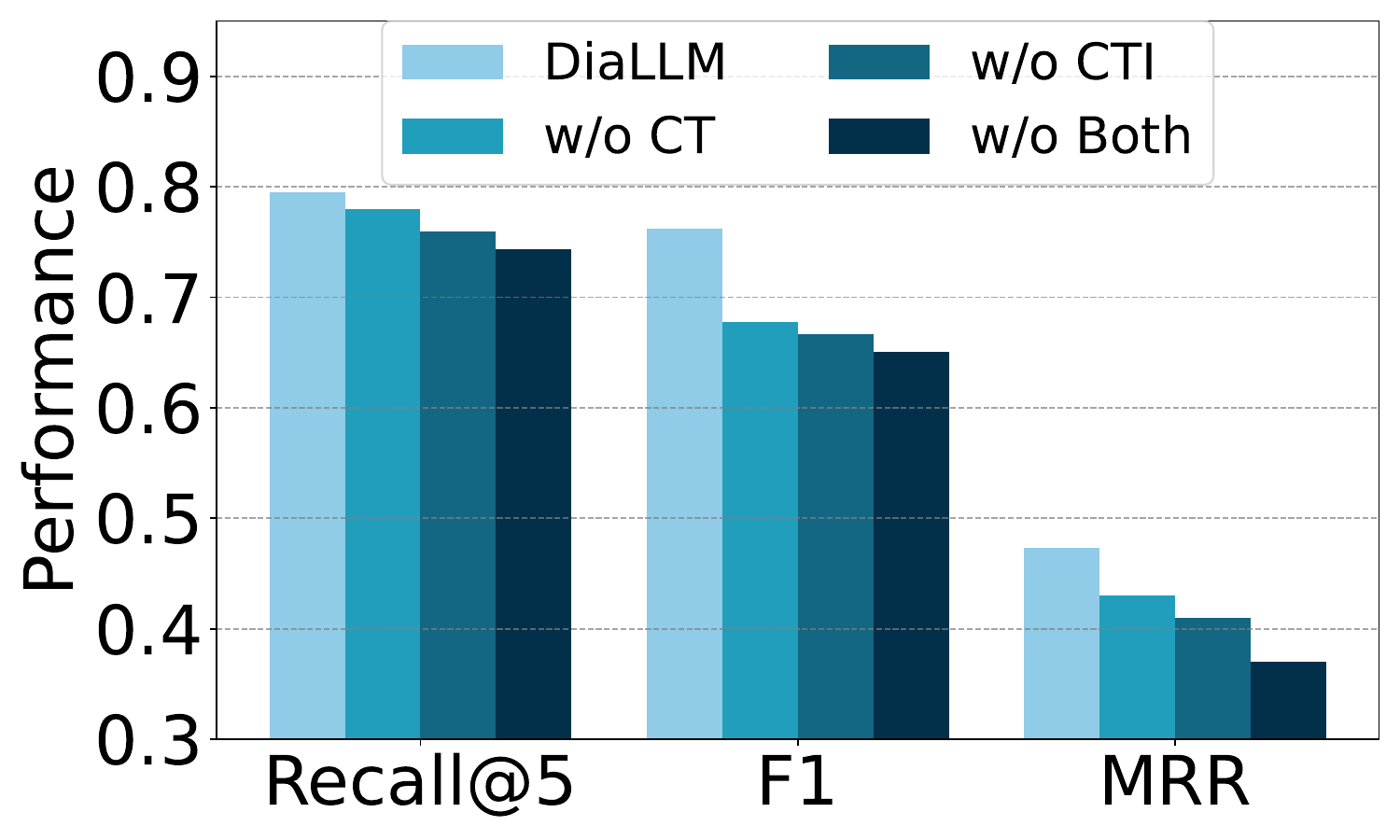}
        \label{fig:subfig1}
    }
    \subfigure[TriNetX-Respiratory Dataset.]{
        \includegraphics[width=0.3\textwidth]{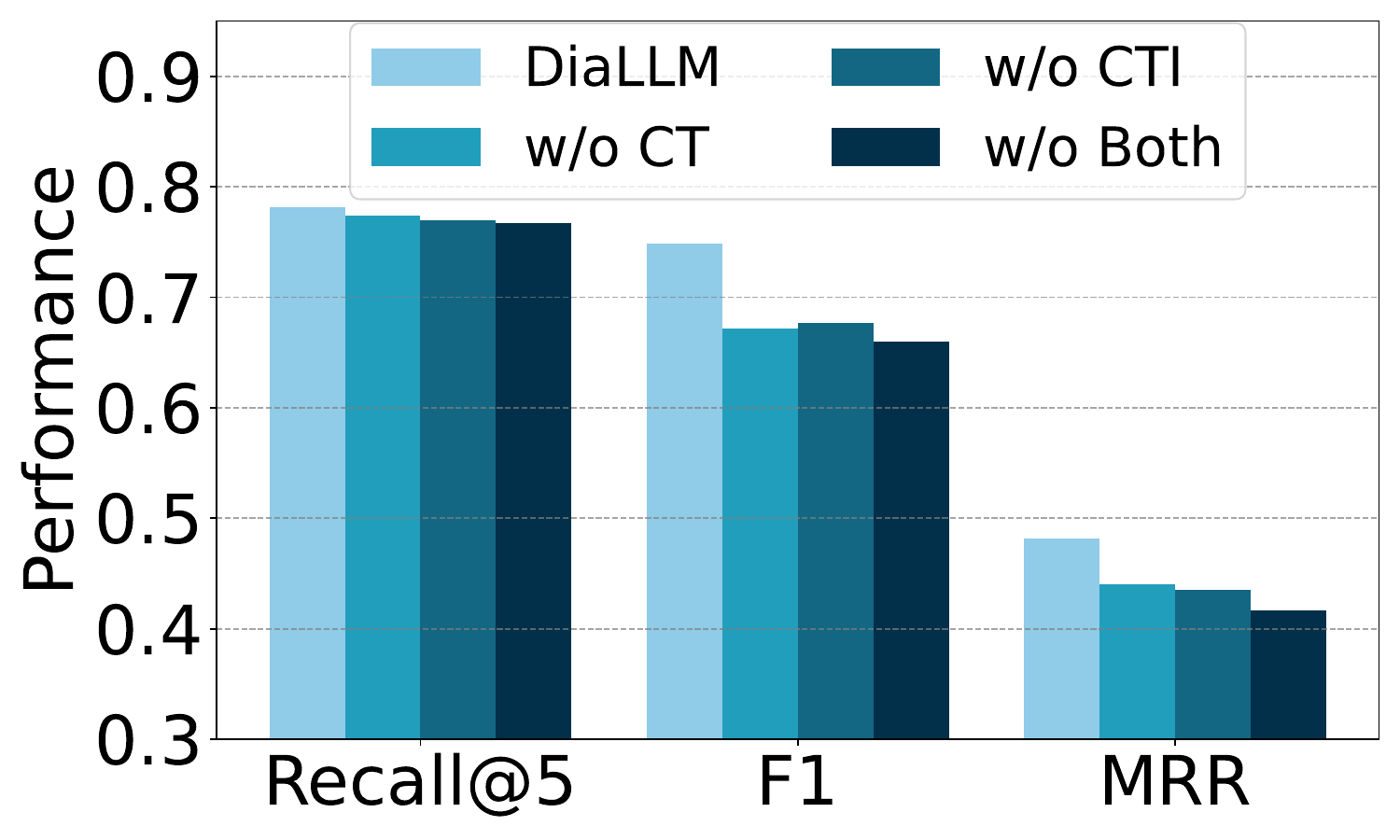}
        \label{fig:subfig2}
    }
    \subfigure[TriNetX-Circulatory Dataset.]{
        \includegraphics[width=0.3\textwidth]{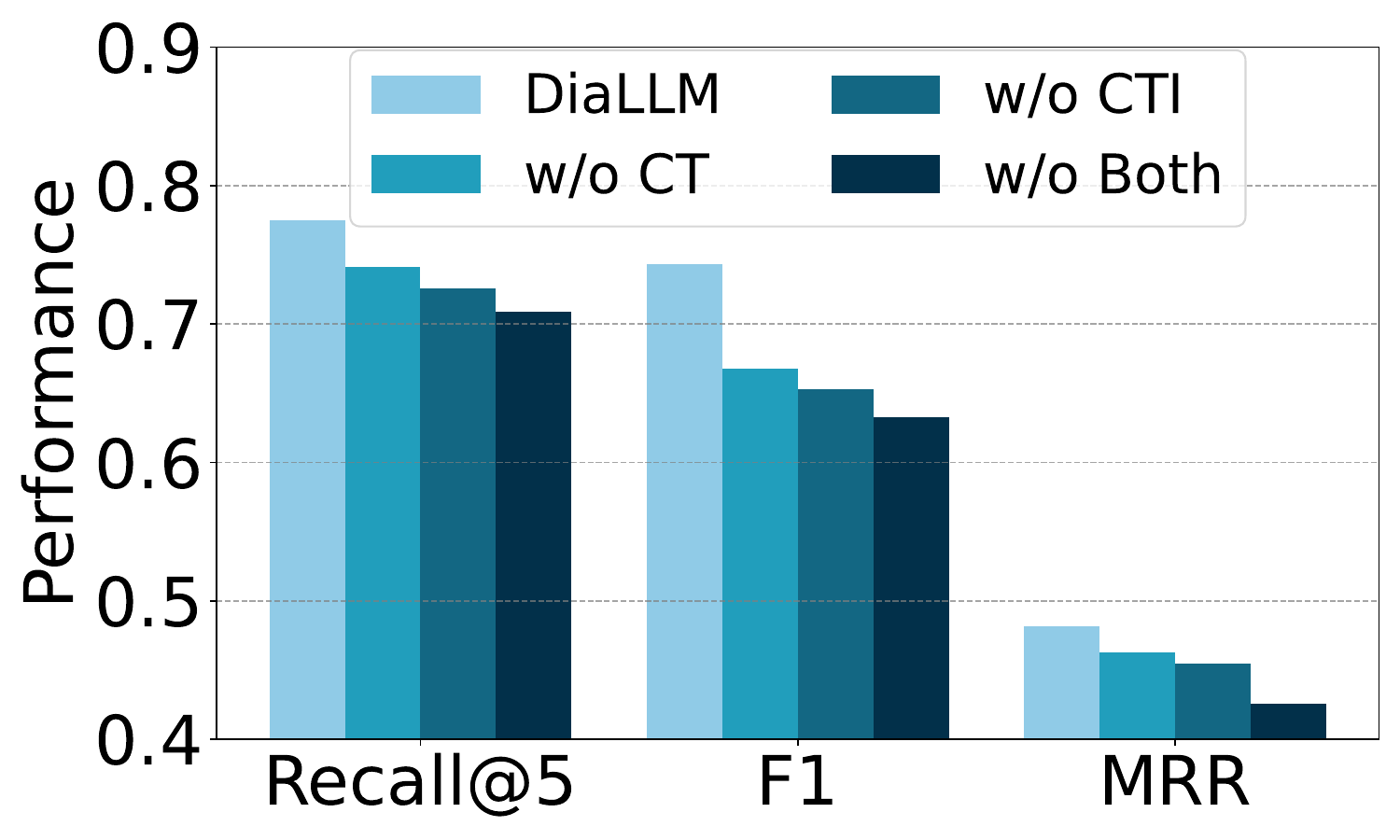}
        \label{fig:subfig3}
    }
    \caption{Ablation Study on Single-turn Dialogue Data.}
    \label{fig:ablation_single}
\end{figure*}

%% file: Figures/ablation/ablation_multi.tex
\begin{figure*}[t]
    \centering
    \subfigure[TriNetX-Metabolic Dataset.]{
        \includegraphics[width=0.3\textwidth]{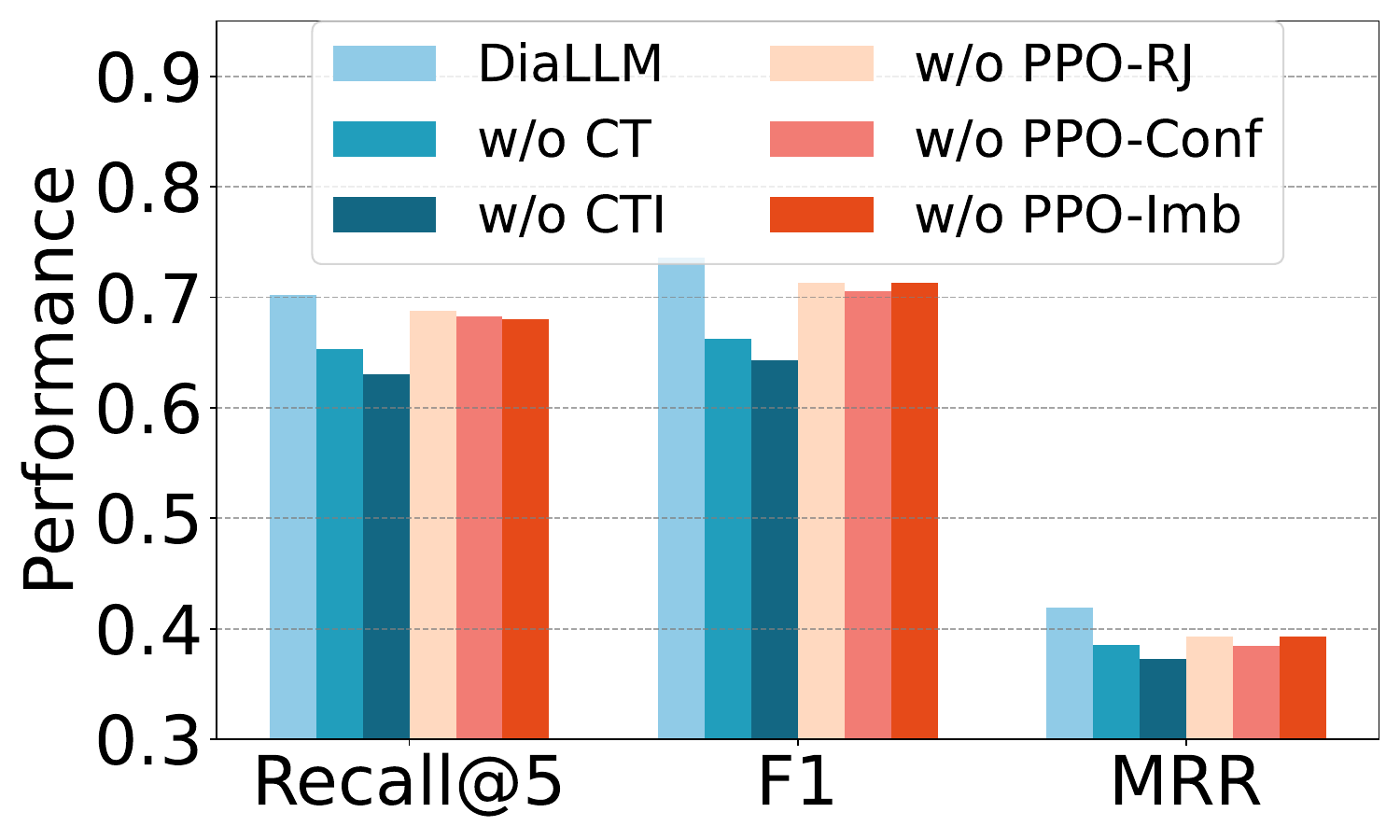}
        \label{fig:subfig1}
    }
    \subfigure[TriNetX-Respiratory Dataset.]{
        \includegraphics[width=0.3\textwidth]{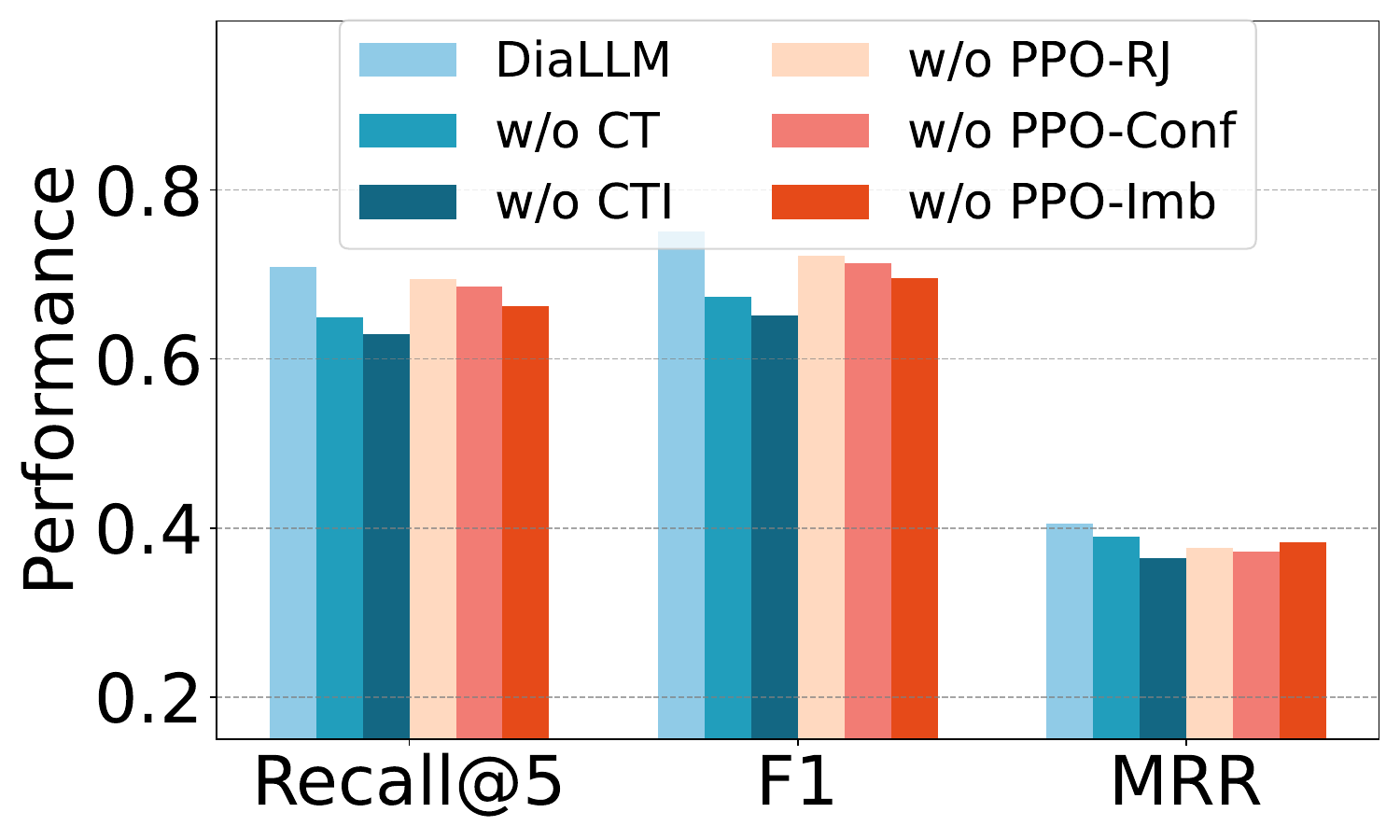}
        \label{fig:subfig2}
    }
    \subfigure[TriNetX-Circulatory Dataset.]{
        \includegraphics[width=0.3\textwidth]{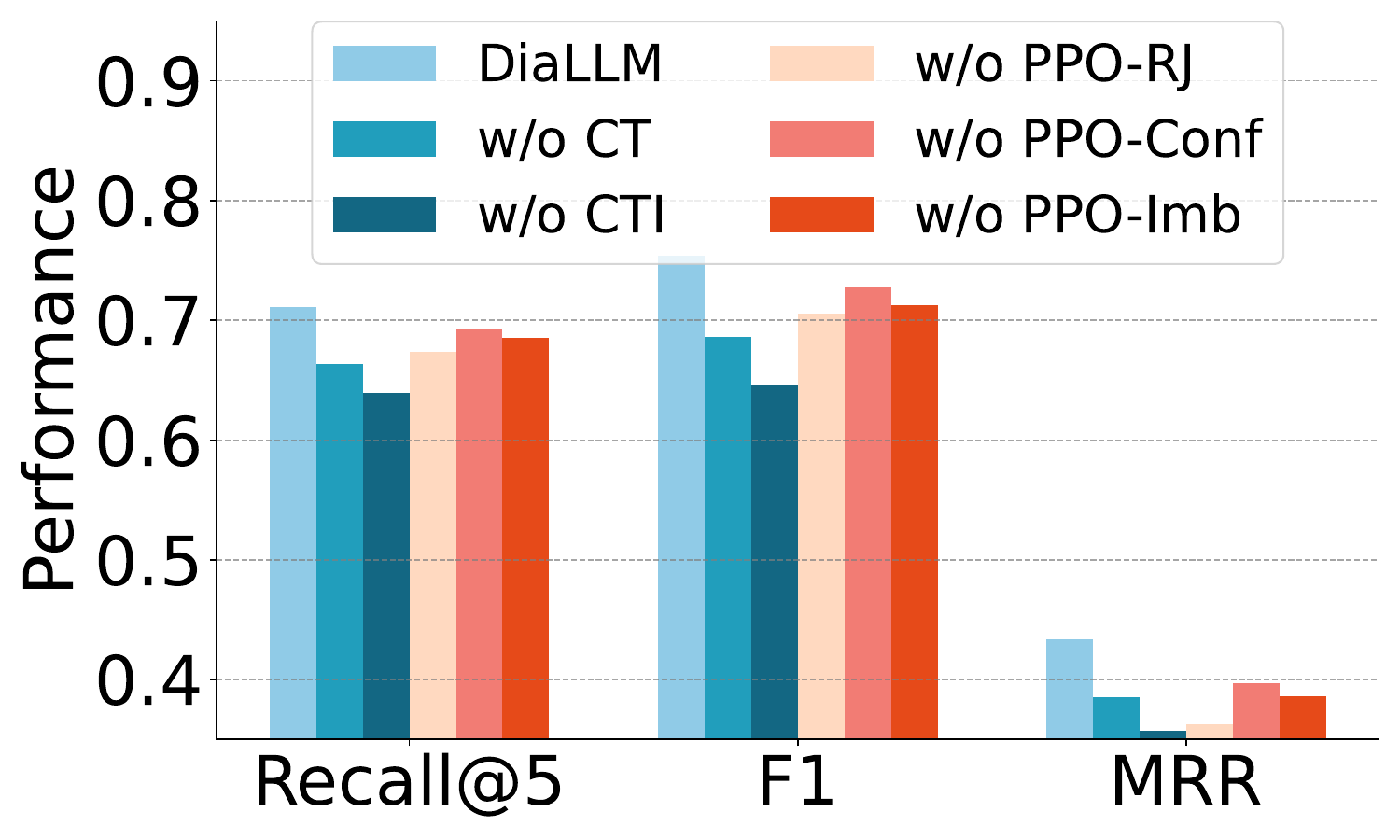}
        \label{fig:subfig3}
    }
    \caption{Ablation Study on Multi-turn Dialogue Data.}
    \label{fig:ablation_multi}

\end{figure*}

%% file: Sections/conclusion.tex
\section{Conclusion}
In this work, we introduced DiaLLM, a clinical dialogue model that enhances medical conversational systems through a combination of clinical-grounded EHR transformation and PPO optimization. 
Experimental results consistently show that DiaLLM outperforms existing models across a variety of metrics, demonstrating its superior prediction ability.
By integrating domain-specific knowledge and a structured decision-making process, DiaLLM offers a significant improvement over both general-purpose and medical-specific LLMs in complex medical dialogues.

%% file: Sections/Limitations.tex
\section*{Limitations}


\noindent \textbf{Automation of CTR Database}
In this work, we recruited three medical students to annotate clinical test code descriptions and clinical value reference ranges based on medical guidelines and publicly available clinical resources (e.g., Mayo Clinic). 
Our constructed Clinical Test Reference data are crucial for diagnosis prediction using EHR data grounded in clinical knowledge.
However, manual annotation is time-consuming and prone to variability. Future work will focus on automating this process to enhance efficiency and consistency.

\noindent \textbf{Clincial Conversational System with Multimodal Data} 
In this work, we focus on diseases that can be solely predicted through EHR data. 
Our approach focuses on EHR data and lab tests with numerical or textual features. With advances in multimodal learning, integrating imaging-based diagnostic tests, such as CT scans, could further enrich the model’s clinical reasoning capabilities. We also leave this extension for future exploration.

\section*{Ethic Statement}

\textbf{Data Collection} This study utilizes publicly available EHR data and de-identified in-hospital records approved for research use. To safeguard patient privacy, we conducted a thorough review to ensure the dataset contains no sensitive information. We adhere to strict ethical standards in data handling and acknowledge the research community’s commitment to maintaining data integrity and privacy.

\noindent 
\textbf{Compensation and Ethical Collaboration with Medical Professionals} 
We collaborated with two senior medical graduates to identify diseases predictable solely from EHR data and to evaluate the constructed CTR. Each was compensated 200 RMB per hour, aligning with local salary standards to ensure fair remuneration for their expertise.

\noindent 
\textbf{Concerning the Trustworthiness of DiaLLM}
While DiaLLM demonstrate strong diagnostic reasoning capabilities, its reliability in real-world clinical settings is not yet fully established. 
Potential risks include hallucinations, biases from training data, and inconsistencies in medical reasoning. To enhance trust, further research is needed to improve model transparency, robustness, and validation through expert evaluation. Developing rigorous verification mechanisms and integrating clinician oversight will be essential for safe deployment in medical practice.

%% file: Sections/Appendix.tex
\section{Appendix}
\input{Tables/Appendix/diabet}

\subsection{Methodology}
\subsubsection{Illustration of EHR data.}
\input{Tables/appendix_table_EHR}
We summarize main elements and corresponding examples in EHR data in Table \ref{tab:element_ehr}. 
EHR data is heterogeneous and can be represented as a sequence of tables.
\subsubsection{Dialogue Data Construction from EHR}
The EHR transformation comprises two components (1) Clinical Code Translation and (2) Clinical Test Interpretation. Examples of our constructed \textit{Clinical Code Translation} data is shown in 
Table \ref{tab:clinical-description}
and examples of \textit{Clinical Test Interpretation} data is shown in \ref{tab:clinical-reference}, respectively.
\input{Tables/appendix_code_mapping}
\input{Tables/lab_code_example}

\subsection{Experiments}

\subsubsection{Statistical Analysis of Datasets.}
We exclude clinical tests with a frequency of less than 10 across the entire dataset. Additionally, we filter out tests for which no test results are available for each patient.
\label{app::statistic}
Statistical analysis of datasets is shown in Table \ref{tab:Statistical_Datasets}.
\input{Tables/Appendix/EHR_statistics}

\subsubsection{Evaluation Metrics} 
\label{sec::evaluation_metric}
The Mean Reciprocal Rank (MRR) is a metric used to evaluate systems that return a ranked list of answers to queries, focusing on the position of the first relevant answer. 
A higher MRR indicates that relevant items tend to appear higher in the ranked list of results.
It is defined as the average of the reciprocal ranks of the first relevant answer for a set of queries.
Formally, MRR is expressed as:

\begin{equation}
\text{MRR} = \frac{1}{|Q|} \sum_{i=1}^{|Q|} \frac{1}{\text{rank}_i}
\end{equation}

\subsubsection{Main Results on NHANES Dataset}
\label{app::NHANES}
Experimental results on the NHANES dataset is shown in Table \ref{tab:NHANES_single_turn}. It can be observed that: (1) Medical-specialized models consistently outperform general foundation models; (2) When adapting foundation models with an MLP classifier trained on extracted embeddings, performance improves drastically; (3) DiaLLM significantly outperforms all baselines, demonstrating its superior ability in diagnosis prediction. These results highlight the effectiveness of DiaLLM’s integration of structured EHR data and reinforcement learning framework, enabling more accurate and clinically relevant predictions compared to both zero-shot LLMs and embedding-based MLP classifiers.

\subsubsection{Ablation Study on NHANES Dataset}
\label{app::aba_nhanes}
\input{Figures/Appendix/diabetes_ablation}
Experimental results on the NHANES dataset is shown in Figure \ref{Fig:ablation_mhanes}.
The NHANES dataset presents a simpler task compared to TriNetX data. The ablation study highlights the critical role of 
clinically grounded data transformation and rewarding modeling for improving diagnosis performance. 

\subsubsection{Case Study}
\label{app:case_study}

\input{Tables/Appendix/case_study_multi}
 The case study example is presented in Table \ref{tab::single_turn_prompt}.

%% file: Tables/Appendix/diabet.tex
\begin{table}[h]
\centering
\resizebox{0.45\textwidth}{!}{
\begin{tabular}{l ccc }
\toprule
\multirow{2}{*}{\textbf{Model}} & \multicolumn{3}{c}{\textbf{NHANES}}  \\  
& Recall@1 & F1 & MRR \\ 
\midrule
Mistralv0.3-7B & $28.60$ & $40.48$ & $37.41$ \\
Llama3.1-8B & $29.70$ & $45.69$ & $47.64$ \\
Qwen2.5-7B & $28.97$ & $44.83$ & $44.38$ \\
\midrule
BioMistral & $36.19$ & $48.25$ & $43.13$ \\
ChatDoctor & $39.83$ & $54.29$ & $46.72$ \\
Meditron-7B & $36.82$ & $52.47$ & $47.77$ \\
Meditron3-8B & $37.69$ & $55.34$ & $49.78$ \\
\midrule
Mistralv0.3-7B$_{mlp}$ & $91.98$ & $92.47$ & $ 97.99$ \\
Llama3.1 8B$_{mlp}$ & $92.15$ & $92.66$ & $ 98.06$ \\
Qwen2.5-7B$_{mlp}$ & $92.07$ & $92.45$ & $98.02$ \\

\midrule
DiaLLMs & \textbf{95.94}  & \textbf{96.07} & \textbf{98.97}  \\ 
\bottomrule
\end{tabular}
}
\caption{Performance Comparison of DiaLLM and Baselines on NHANES Dataset.}
\label{tab:NHANES_single_turn}
\end{table}

%% file: Tables/appendix_table_EHR.tex
\begin{table*}[h]
    \centering
\resizebox{0.9\textwidth}{!}{ 
    \begin{tabular}{cccc}
\toprule
\toprule
        \textbf{EHR Data} & \textbf{Description} & 
        \textbf{Event} &
        \textbf{Visit Time} \\
\midrule
Demographics 
& General characteristics of patients 
& Age, gender, ethnicity/race &\\
\midrule
Symptom & \makecell[c]{Patient symptoms are cataloged in the \\ICD system under codes R00-R99.} &  \makecell[c]{ICD-10-CM R10.12: Left\\ upper quadrant pain}& 2019-02-15\\
\midrule
Vital Signs 
&\makecell[c]{Medical signs indicating \\the status of the body’s
\\vital functions }
& \makecell[c]{
LOINC 8462-4 82 mm[Hg] \\ 
LOINC 29463-7, 84 [lb$\_$av]\\
LOINC 8480-6, 124 mm[Hg]
}
&
\makecell[c]{
2019-02-15 \\ 
2019-02-19 \\
2019-02-22 
}
\\
\midrule

\makecell[c]{Laboratory \\Results}
& \makecell[c]{Medical examination results, \\generate are organized \\in a code format} 
& \makecell[c]{
LOINC 73578-7,8 mmol/L, \\
LOINC 6768-6 48 U/L, \\
LOINC 19261-7 Negative} 
&
\makecell[c]{
2019-02-15 \\ 
2019-02-19 \\
2019-02-22 
}
\\
\midrule

Diagnostic 
& \makecell[c]{Codes representing diseases \\ and related health
problems \\ (e.g., ICD-9 and ICD-10) }
& \makecell[c]{ICD-10-CM Q65: Congenital \\deformities of hip, \\ICD-10-CM M25.552: Pain in left hip, 
\\ICD-10-CM R06.02: Shortness of breath} 
&2019-02-22 \\
\bottomrule
\bottomrule

\end{tabular}
}
\caption{Description of key components and corresponding examples in EHR data.}
\label{tab:element_ehr}

\end{table*}

%% file: Tables/appendix_code_mapping.tex
\begin{table*}[ht]
\centering
\resizebox{0.9\textwidth}{!}{
\begin{tabular}{c|c}
\hline
\textbf{LOINC Code} & \textbf{Description} \\
\hline
100091-8 & Trypanosoma cruzi Ab [Units/volume] in Serum by Immunoassay \\
100092-6 & Trypanosoma cruzi Ab bands panel - Serum by Immunoblot \\
100093-4 & Trypanosoma cruzi 15-16kD IgG Ab [Presence] in Serum by Immunoblot \\
100094-2 & Trypanosoma cruzi 21-22kD IgG Ab [Presence] in Serum by Immunoblot \\
100095-9 & Trypanosoma cruzi 27-28kD IgG Ab [Presence] in Serum by Immunoblot \\
100096-7 & Trypanosoma cruzi 42kD IgG Ab [Presence] in Serum by Immunoblot \\
100097-5 & Trypanosoma cruzi 45-47kD IgG Ab [Presence] in Serum by Immunoblot \\
100098-3 & Trypanosoma cruzi 120-200kD IgG Ab [Presence] in Serum by Immunoblot \\
100099-1 & Trypanosoma cruzi 160kD IgG Ab [Presence] in Serum by Immunoblot \\
1001-7 & DBG Ab [Presence] in Serum or Plasma from Donor \\
10010-7 & R' wave amplitude in lead AVF \\
100100-7 & Fasciola sp IgG Ab [Presence] in Serum by Immunoassay \\
100101-5 & Fasciola sp 8-9kD IgG Ab [Presence] in Serum by Immunoblot \\
100102-3 & Fasciola sp 27-28kD IgG Ab [Presence] in Serum by Immunoblot \\
\hline
\end{tabular}
}
\caption{The Mapping between LOINC Codes and their Corresponding Clinical Descriptions.}
\label{tab:clinical-description}
\end{table*}

%% file: Tables/lab_code_example.tex
\begin{table*}[h!]
\centering
\renewcommand{\arraystretch}{1.2}
\setlength{\tabcolsep}{6pt} 
\begin{adjustbox}{width=0.9\textwidth} %
\begin{tabular}{ c l c c c c c}
\hline
\textbf{Lab Code} & \textbf{Test Name}& \textbf{Age Range} & \textbf{Gender} & \textbf{Reference Range} & \textbf{Unit}& \textbf{Critical Value}                        \\ \hline
2823-3            & Potassium Serum                               & 1–18 years         &     Any            & 3.4–4.7                 & mEq/L                 &  -                                            \\ \hline
2823-3            & Potassium Serum                               &Any $\geq$18 years          &  Any               & 3.5–5.2                 & mEq/L                 &       -                                       \\ \hline
17861-6           & Total Calcium                                 & $<$1 year            &         Any        & 8.7–11.0               & mg/dL                 &   -                                           \\ \hline
17861-6           & Total Calcium                                 & 1–17 years         &    Any             & 9.3–10.6               & mg/dL                 &   -                                           \\ \hline
17861-6           & Total Calcium                                 & 18–59 years        &     Any            & 8.6–10.0               & mg/dL                 &   -                                           \\ \hline
17861-6           & Total Calcium                                 & $\geq$60 years          &          Any       & 8.8–10.2               & mg/dL                 &     -                                         \\ \hline
33914-3 & \makecell[l]{eGFR (Estimated \\ Glomerular Filtration Rate)} & Any& Any& $>$90& mL/min/1.73m²         &   -                                           \\ \hline
33914-3           & \makecell[l]{eGFR (Estimated \\ Glomerular Filtration Rate)} & Any & Any& None&mL/min/1.73m²& \makecell[l]{Mildly low: 60–89  } \\ \hline
33914-3           & \makecell[l]{eGFR (Estimated \\ Glomerular Filtration Rate)} & Any                & Any             & None                   &   mL/min/1.73m² & \makecell[l]{Moderate to severe \\ kidney disease: \\ 30–59 } \\ \hline
33914-3           & \makecell[l]{eGFR (Estimated \\ Glomerular Filtration Rate)} & Any                & Any             & None                   &  mL/min/1.73m² & \makecell[l]{Kidney failure: \\ $<$15 } \\ \hline
\end{tabular}
\end{adjustbox}
\caption{Examples of the Manually Constructed Clinical Test Interpretation Data.}
\label{tab:clinical-reference}
\end{table*}

%% file: Tables/Appendix/EHR_statistics.tex
\begin{table}[H]
\centering
\resizebox{0.45\textwidth}{!}{ 
\begin{tabular}{cp{1.2cm}p{1.4cm}p{1.3cm}p{1.2cm}}
\hline
& \makecell[c]{TriNetX-\\Metabolic} & \makecell[c]{TriNetX-\\Respiratory} & \makecell[c]{TriNetX-\\Circulatory} & \makecell[c]{NHANES} \\
\hline
$\#$Samples& 15844 & 10068 & 6747 &  8897 \\
$\#$Diagnosis & 49 &  47& 49 & 4\\
$\#$Clinical Test & 272 & 301 & 217 & 45  \\
\hline
\end{tabular}
}
\caption{Statistical Analysis of Datasets.}
\label{tab:Statistical_Datasets}
\end{table}

%% file: Figures/Appendix/diabetes_ablation.tex
\begin{figure}
\centering
\includegraphics[width=0.45\textwidth]{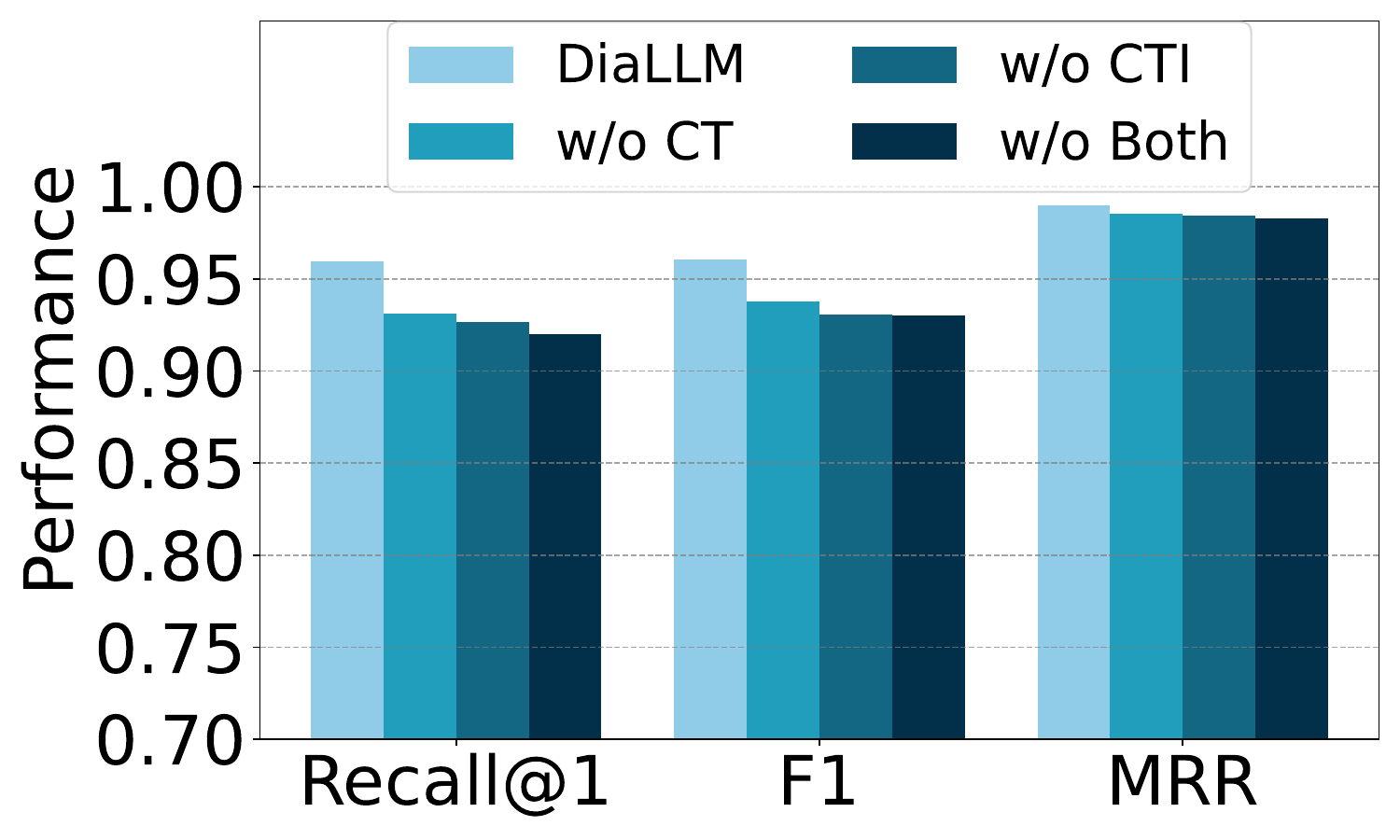}
\caption{ Ablation Study on NHANES Dataset.}
\label{Fig:ablation_mhanes}
\end{figure}

%% file: Tables/Appendix/case_study_multi.tex
\begin{table*}[htbp]
\centering

\resizebox{0.85\textwidth}{!}{
\begin{tabular}{p{15cm}}  
\toprule
\textbf{A Sub-piece of Initial EHR Data:} \\
\colorbox{case_1}{Demographic Information: age: 30, gender: female, race: white}\\ 
\colorbox{case_2}{Symptom Code}: R10.2, R10.32\\
\colorbox{pink}{Lab Test Code}:
2028-9, 2075-0,  2951-2, 3040-3, 33037-3, 48642-3, 777-3, 
1742-6, 2028-9, 2075-0
\\
\colorbox{pink}{Lab Results Values}:
31.0, 105.0, 83.0, 139.0,7.0, 99.0, 283.0, 27.0, \\
\colorbox{case_3}{Vital Sign Code}:  8302-2, 8310-5, 8462-4, 8480-6, 9279-1 \\
\colorbox{case_3}{Vital Sign Values}: 
 62.0, 98.0, 70.7, 119.7, 98.7\\

Diagnosis Codes: E03.9, K80.62, K81.0, N80.9, K80.20, K80.10 \\

\midrule
\midrule
\textbf{Prompts:} \\
\colorbox{case_1}{Given the patient demographic information}: (age: 30, gender: female, race: white).

\colorbox{case_2}{Patient shows symptom of}: (1) Pelvic and perineal pain; (2) Right lower quadrant pain.

\colorbox{pink}{These lab tests show normal results}:
(1) Carbon dioxide, totalin Serum or Plasma, 
(2) Chloridein Serum or Plasma, 
(3) Glucosein Serum or Plasma, (4) Sodiumin Serum or Plasma,
(5) Lipasein Serum or Plasma,
(6) Glomerular filtration rate/1.73 sq M predicted among non-blacksin Serum or Plasma by Creatinine-based formula (MDRD).

\colorbox{pink}{These lab tests show abnormal results}:
(1) Plateletsin Blood by Automated count.

\colorbox{case_3}{These vital sign show normal Results}:
(1) Temperature,
(2) Blood Pressure (Systolic),
(3) Blood Pressure (Diastolic),
(4) Respiratory Rate,

\colorbox{case_3}{These vital sign show abnormal values}:
(1) Heart Rate

What are your recommended diagnosis? If you need further information, what are your recommended clinical test for further decision making?\\

\midrule
\textbf{DiaLLM}\\
I recommend you to take (1) Alanine aminotransferase in Serum or Plasma; (2) Carbon dioxide in Serum or Plasma, (3) Chloride in Serum or Plasma. \\

\textbf{Prompts (User)}: \\
My Alanine aminotransferase in Serum or Plasma results is 27.0 $U/L$ , Carbon dioxide in Serum or Plasma is 83.0 $mEq/L$, Chloride in Serum or Plasma is 139.0 $mEq/L$.\\

\textbf{DiaLLM}:\\

These lab test shows normal results: 
(1) Alanine aminotransferase in Serum or Plasma results;

These lab test shows abnormal results:
(1) Carbon dioxide in Serum or Plasma,
(2) Chloride in Serum or Plasma.
I further recommend you to take 
(1) Glucose in Serum or Plasma;
(2) Sodium in Serum or Plasma;

\textbf{Prompts (User)}: \\
My Glucose in Serum or Plasma is 7.0 $mmol/L$,
(2) Sodium in Serum or Plasma 99.0 $mmol/L$.

\textbf{DiaLLM}:\\
These lab test shows abnormal results:
(1) Glucose in Serum or Plasma;
(2) Sodium in Serum or Plasma.
I recommend the following possible diagnosis:
(1) disorders of thyroid gland, 
(2) disorders of gallbladder, biliary tract and pancreas,
(3) Noninflammatory disorders of female genital tract.
\\
\midrule
\textbf{BioMistral}\\
I recommend the following possible diagnosis:
(1) Chlamydia trachomatis infection,
(2) Pelvic inflammatory disease,
(3) Pelvic pain
\\
\midrule
\textbf{Meditron-7B} \\
I recommend the following possible diagnosis:
(1) Pelvic inflammatory disease,
(2) Acute appendicitis,
(3) Ovarian torsion,
(4) Ectop \\
\midrule
\textbf{Meditron-8B} \\
I recommend the following possible diagnosis:
(1) Pelvic inflammatory disease,
(2) Chlamydia trachomatis infection,
(3) Pelvic inflammatory disease with chlamydia trach. \\
\midrule

\textbf{Chatdoctor}\\
I recommend the following possible diagnosis:
(1) pelvic inflammatory disease,
(2) tuberculosis,
(4) acute pancreat.\\
\midrule
\bottomrule
\end{tabular}
}
\caption{Sample Responses for Case Study.}
\label{tab::single_turn_prompt}
\end{table*}